\documentclass[10pt,twocolumn,twoside]{IEEEtran}
\usepackage{amsfonts}
\usepackage{graphicx}
\usepackage{epstopdf}
\usepackage{subcaption}
\usepackage{multirow}
\usepackage{amsmath}
\usepackage{amssymb}
\usepackage{bm}
\usepackage{float}
\usepackage{url}
\usepackage{booktabs} 
\usepackage{balance} 
\usepackage{array}
\usepackage{xcolor}
\newcolumntype{L}[1]{>{\raggedright\let\newline\\\arraybackslash\hspace{0pt}}m{#1}}

\newcommand{\eat}[1]{} 
\begin{document}

\title{Dual Attention on Pyramid Feature Maps for Image Captioning}
 
\author{Litao~Yu, 
	Jian~Zhang,~\IEEEmembership{Senior~Member,~IEEE,}
	Qiang~Wu,~\IEEEmembership{Senior~Member,~IEEE}
\thanks{Litao Yu (litao.yu@uts.edu.au), Jian Zhang (jian.zhang@uts.edu.au) and Qiang Wu (qiang.wu@uts.edu.au) are with Global Big Data Technologies Centre, University of Technology Sydney, Ultimo, 2007, NSW, Australia.}
 \thanks{Manuscript received August 14, 2020. Corresponding author: Jian Zhang}
}
 
\maketitle
 
\markboth{IEEE Transactions on Multimedia}%
 {Yu \MakeLowercase{\textit{et al.}}:Dual Attention on Pyramid Feature Maps for Image Captioning}

\begin{abstract}

Generating natural sentences from images is a fundamental learning task for visual-semantic understanding in multimedia. In this paper, we propose to apply dual attention on pyramid image feature maps to fully explore the visual-semantic correlations and improve the quality of generated sentences. Specifically, with the full consideration of the contextual information provided by the hidden state of the RNN controller, the pyramid attention can better localize the visually indicative and semantically consistent regions in images. On the other hand, the contextual information can help re-calibrate the importance of feature components by learning the channel-wise dependencies, to improve the discriminative power of visual features for better content description. We conducted comprehensive experiments on three well-known datasets: Flickr8K, Flickr30K and MS COCO, which achieved impressive results in generating descriptive and smooth natural sentences from images. Using either convolution visual features or more informative bottom-up attention features, the composite model can boost the performance of image-to-sentence translation, with a limited computational resource overhead. The proposed pyramid attention and dual attention methods are highly modular, which can be inserted into various image captioning modules to further improve the performance.

\end{abstract}

\begin{IEEEkeywords}
Image Captioning; Dual Attention; Pyramid Attention.
\end{IEEEkeywords}

\section{Introduction}

The image captioning task is to apply machine learning methods to reveal the image contents and generate descriptive natural sentences, which combines visual understanding in computer vision and machine translation in natural language processing. The target of image captioning is to bridge the semantic gap between visual feature descriptors and human languages. Usually, an image captioning framework consists of two sub-models, an encoder that extracts the visual features from images, and a decoder that translates the visual properties to natural sentences. The encoder is essentially a pre-trained convolutional neural network (CNN), and the decoder is mainly a recurrent neural network (RNN) that controls the word-flow in the sentences. Thus, the challenges of image captioning are two-fold: first, as a visual understanding task, the feature representation should be discriminative enough to determine what visual properties to describe; second, the language model should be able to generate sentences that accurately describe the semantics in images. The second requirement differs from cross-modal retrieval tasks \cite{MM17:ACMR,TIP:LDB,TPAMI:LSRH}, which do not need to assemble words and form smooth sentences.

\begin{figure}[t]
\centering
    \begin{minipage}{0.23\textwidth}
        \includegraphics[width=1\textwidth]{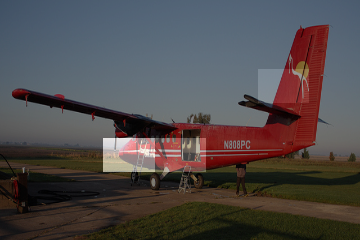}
        \subcaption{Single-scale attention}
	   \label{FIG:SA}
    \end{minipage}
    ~ 
    \begin{minipage}{0.23\textwidth}
        \includegraphics[width=1\textwidth]{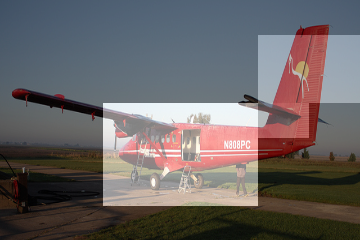}
        \subcaption{Multi-scale attention}
	   \label{FIG:MA}
    \end{minipage}
        
\caption{Single-scale {\em vs} multi-scale attention maps. Caption: a small \underline{{\bf red airplane}} sitting on top of an airport tarmac.}
\label{FIG:ILLUSTRATION}
\end{figure}

From the perspective of instance-based visual understanding, image captioning is a many-to-many learning task, i.e., both image inputs and sentence outputs contain multiple instances, and they should be well correlated for sentence generation. Inferring with a unique global visual feature vector is not a good option to describe multiple instances in an image, because the many-to-many correlations between the two modalities are usually lost, thus aggregating local visual descriptors is more suitable for word generation in a salient region at different time steps in a sentence. With the help of the partial contextual information, such aggregation is feasible via the control of the RNN module in the decoder, and these salient regions can be localized in accordance with the temporal order of the natural sentence, which is called attention as a mechanism for visual feature selection. The temporal attention model is specifically designed for the task of machine translation \cite{TPAMI:DA,JTNMT:ABJT}. As a kind of image-to-sentence translation, attention models have also been used in learning-based image captioning. In \cite{ICML15:SAT}, the authors introduced two attention-based image captioning models under a common framework: a “soft” deterministic attention and a “hard” stochastic attention, respectively. The former model is trained by standard back-propagation method while the latter one is trained in a reinforcement learning approach. Following this work, various image captioning models have been proposed to further boost the captioning performance \cite{CVPR16:SA,CVPR17:SCA,CVPR17:BA,CVPR18:BUTD,CVPR19:LBPF}. While encouraging results on public datasets have been reported, the performance gains obtained from the latest proposed image captioning models mainly rely on additional deep learning models such as attribute learning \cite{CVPR17:BA} and region proposal network \cite{CVPR18:BUTD}, to get fine-tuned semantic information. Such conditions are not always satisfied either because the lack of label annotation or the computational resource limitation.

We rethink the importance of visual feature representation for better attentions in a single inference model, to improve the performance of image captioning. In the decoder of an attention-based image captioning model, the RNN controls a small region of the input image to gaze at a time step. Such a region is represented by a feature vector aggregated by a soft-max function or a hard mask function in a salient area of an input image. However, the objects usually render at multi-scales, so it is inaccurate if the attention module just looks at one position at a single scale. Consequently, the single-scale spatial attention may lead to the mismatched visual-semantic relationship because of the inconspicuous classes. As is illustrated in Figure \ref{FIG:ILLUSTRATION}, the {\em red airplane} should be represented in a larger receptive field in the image, and such kind of 2D areas usually cannot be well described via a softmax function over the whole convolutional feature maps. In \cite{CVPR18:BUTD}, the authors proposed to use a pre-trained object detector to estimate the bounding boxes of the entities, forming a bottom-up feature representation. Such an attention mechanism becomes a very good start point for improving the semantically correct mapping for image captioning \cite{ICCV19:AOA,ICCV19:HIP,CVPR19:LBPF}.

In this paper, we propose to apply a pyramid attention module on visual image feature maps to form multi-scale representations and generate richer attentions. When using the convolutional features without the help of the auxiliary object detector, the attention can still capture the visual properties accurately to facilitate the image description. When using the bottom-up image features, the pyramid attention yields semantic hierarchies. In either case, the pyramid feature representations can outperform the single-scale feature maps. Furthermore, we employ a dual attention module from both spatial and channel perspectives in the decoder part. The spatial attention is conducted at a local image area when given the contextual information of a hidden state of RNN, while the channel attention is to re-calibrate the feature components from a different view of feature maps. With such settings, both the spatial attention on pyramid feature maps and the dual attention on the visual feature representations can improve the image feature representations separately. When combining the two modules to form a unified image captioning framework, our proposed method is able to achieve very competitive performance as a single image captioning model in terms of BLEU \cite{ACL02:BLEU}, METEOR \cite{ACL05:METEOR}, ROUGE-L \cite{ACL04:ROUGE}, and CIDEr-D \cite{CVPR15:CIDER}, on three publicly available datasets: Flickr8K \cite{JAIR:FLICKR8K}, Flickr30K \cite{ACL14:FLICKR30K} and MS COCO \cite{ECCV14:COCO}.

The rest of the paper is organized as follows. Section \ref{SEC:RELATED_WORK} introduces related work. Section \ref{SEC:METHOD} elaborates the proposed dual attention on pyramid feature maps for image captioning. Experimental results and analysis are presented in Section \ref{SEC:EXP}. Finally, Section \ref{SEC:CONCLUSION} concludes the paper.

\section{Related work}
\label{SEC:RELATED_WORK}

\subsection{Image captioning}

Inspired by the successful use of convolutional neural networks (CNNs) in computer vision and recurrent neural networks (RNNs) in natural langue processing, a large number of deep learning-based visual captioning methods have been proposed in recent years. The pioneering work using the encoder-decoder structure to generate natural sentences is proposed by Vinyals et al. \cite{CVPR15:IMCAP}, which is now still the most widely used captioning structure. In \cite{CVPR18:CONVIMCAP}, Aneja et al. proposed an alternative method using the 1D convolutional decoder to generate sentences, which has the comparable performance to the RNN based decoder, while has a faster training time per number of parameters. The research targets of image captioning can be categorized as the following three aspects: (1) how to accurately describe the image contents \cite{CVPR19:LBPF,CVPR17:BA}, (2) how to improve the training strategy \cite{CVPR17:SPIDER}, and (3) how to evaluate the captioning model \cite{CVPR18:EIC}. In our work, we mainly focus on how to prepare better visual feature representations to improve captioning quality. A related visual understanding task is video captioning, which also aims to generate natural sentences by considering the dynamic properties \cite{TMM:DRP,TMM:STAT}.

\subsection{Feature pyramid}

In image processing, feature pyramid has been heavily used in the era of hand-engineered features, while it is still a common scheme in deep learning. The advantage of the pyramid feature is that the model can ``observe'' receptive fields at multi-scales to localize and recognize the target patterns in a hierarchical manner. For example, in semantic segmentation, pyramid features can assist the dense prediction at pixel-level by considering the contextual visual cues \cite{CVPR17:PSPNET,BMVC18:PAN}. In \cite{CVPR17:FPN}, pyramid strategy combines the low-resolution but semantically strong features with high-resolution but semantically weak features for object detection. Inspired by these successful use of pyramid features, the model proposed in this paper also benefit from this strategy to prepare better visual feature representations thus generate accurate image descriptions.

\subsection{Attention models}

Attention model has been widely used in many machine learning tasks including machine translation \cite{TPAMI:DA,JTNMT:ABJT}, image classification \cite{CVPR18:SE} and semantic segmentation \cite{CVPR19:DAN}. The general idea of attention mechanism is to apply gated weights to enhance or suppress feature components. In \cite{CVPR18:SE}, the authors proposed a squeeze-and-excitation module to enhance the network from the perspective of the channel-wise relationship. The motivation behind this is to explicitly model channel-interdependencies within the module by selectively enhancing useful features and suppressing useless ones. In video-language analysis tasks, attention models have been extensively used to improve the semantic consistency \cite{TMM:VIDCAP}. By co-attending the temporal information in video frames and the sentence descriptions, the VQA model can achieve better performance even without the use of RNNs \cite{AAAI19:PSAC}. In image captioning, various attention methods have been proposed to improve the sentence generation \cite{CVPR17:SCA,CVPR18:BUTD,CVPR16:SA,MM17:AAC}. Specifically, \cite{CVPR18:BUTD} uses a pre-trained object detector to natually form a bottom-up attention, so the 2D visual patterns in images can be accurately localized. In machine translation, the recent proposed transformer model \cite{NIPS17:TRANSFORMER} can well capture the contextual dependencies in natural languages, which was then applied to the attention on attention (AoA) model for image captioning \cite{ICCV19:AOA}. Inspired by the above attention models, our proposed image captioning model applies attention mechanism mainly from the perspective of visual feature representations, to fully explore the visual-semantic correlations between image feature maps and natural sentences.

\section{Method}
\label{SEC:METHOD}

In this section, we first review the general framework of spatial attention-based image captioning model, then introduce the proposed learning methods. Specifically, we detail two improvements in both encoder and decoder, respectively, and explain why they can further boost the captioning performance.  

\subsection{Review of spatial attention for image captioning}
\label{SEC:REVIEW}

The basic deep learning-based image captioning approach generally feeds the image into a CNN for encoding, then runs this encoding into an RNN decoder to generate an output sentence. The model backpropagates based on the error of the output sentence compared with the groud truth sentence calculated by a loss function like cross-entropy/maximum likelihood \cite{CVPR15:IMCAP}. The performance of sentence generation can be further optimized via policy gradient when given a reward (e.g., CIDEr score \cite{CVPR15:CIDER}) of the generated sentence \cite{CVPR17:SPIDER}. However, translating from a single image feature vector confuses the multiple visual patterns that correspond to the words in the predicted sentences, because the spatial dependencies in the 2D feature maps are totally lost. To solve this problem, spatial attention method tries to build the correlation between a specific 2D region in an image and a word (or phrase) in a sentence to implement more accurate mappings.

The image captioning model takes an image $I$ and generates a caption $Y$. $Y$ is encoded as a sequence of $K$ words:
\begin{equation} \label{EQ:Y}
Y = \{\mathbf{y}_t \in\mathbb{R}^K | t=1,\ldots,T \},
\end{equation}
where $\mathbf{y}_t$ is the word probability vector at the time step $t$, $K$ is the vocabulary size and $T$ is the length of the generated sentence.

A pre-trained CNN model is usually used as an encoder for image $I$. For spatial attention models, we would like to keep the spatial information of the image, so the final convolutional output is with the shape $w\times h\times d$, where $w$, $h$ and $d$ are the width, the height and the number of channels of the feature map, respectively. The output of the encoder is a tensor with the shape $w\times h\times d$, which can be conveniently reshaped to a feature map $\mathbf{V}\in\mathbb{R}^{L\times d}$, where $L=w\times h$. $\mathbf{V}$ can be considered as a set of feature vectors $\{\mathbf{V}_i | i=1,\ldots,L\}$, i.e., each row of $\mathbf{V}$ is a $d$-dimensional feature vector describing a small 2D region in image $I$.

The spatial attention in the decoder aims to obtain a context vector $\mathbf{v}^{(s)}_t\in\mathbb{R}^d$, which is weighted by a spatial attention vector $\alpha=[\alpha_1,\ldots,\alpha_L]\in\mathbb{R}^L$ as follows:
\begin{equation}\label{EQ:SA}
\mathbf{v}^{(s)}_t = \frac{1}{L}\sum\limits_{i=1}^{L}\alpha_i\mathbf{V}_i.
\end{equation}

In image captioning, the spatial attention vector $\alpha$ is determined by the contextual semantics of the sentence. Specifically, the temporal dependencies are mainly sketched by the hidden state of an RNN model. Following the common practise, we use Long Short-Term Memory (LSTM) \cite{NC:LSTM} as the RNN controller in the decoder. The LSTM learns the sequential state of a certain cell by using several non-linear mappings in the state-to-state and input-to-state transitions. For image captioning, the spatial attention vector $\alpha$ is computed from a hidden RNN state $\mathbf{h}_{t-1}$ at a time step $t-1$ and the feature map $\mathbf{V}$:
\begin{align} \label{EQ:SPATIAL_WEIGHT}
\mathbf{a} &= \tanh((\mathbf{V} \mathbf{W}_s +\mathbf{b}_s) \oplus \mathbf{h}_{t-1}\mathbf{W}_{hs}), \nonumber \\ 
\alpha &= \text{softmax} (\mathbf{a} \mathbf{W}_a+\mathbf{b}_a),
\end{align}
where $\mathbf{a}$ is the spatial score vector, which controls the location of semantically salient areas. $\mathbf{W}_s$, $\mathbf{W}_{hs}$ and $\mathbf{W}_a$ are mapping matrices that project the feature maps and hidden state to the same dimension, $\mathbf{b}_s$ and $\mathbf{b}_a$ are bias vectors, and $\oplus$ is the broadcast adding operator, respectively. 

From the work-flow of the spatial attention-based image captioning, we have two important observations as follows:
\begin{itemize}
\item At each time step, the hidden state $\mathbf{h}_{t-1}$ of LSTM controls the gaze at only a specific grid, which approximates to a one-hot mask, rather in a larger receptive field of the input image $I$ (see the softmax operation in Eq. (\ref{EQ:SPATIAL_WEIGHT})). Such a setting essentially hurts the mapping from the visual properties to the word in the generated sentence, because a small grid is usually insufficient to describe the object or stuff identity, and it is unusual to compute less peaky weights to attend larger 2D receptive fields. Furthermore, if the spatial attention gives a ``bad'' position of the image region, the decoder is very likely to mismatch the visual-semantic relations. 
\item The hidden state of LSTM essetntially provides the contextual information in captioning. It does not only focuse on ``where to gaze'' in the whole image at each time step, but can also be extended to control the channel dependencies in the encoded visual features. Thus, $\mathbf{h}_{t-1}$ should be further explored to select the most useful components of the feature vector and suppress the less useful ones when given the contextual information of the sentence. 

\end{itemize}

To summarize the above observations, the visual-semantic inconsistencies are related to the scales of attention. At the same time, the feature map in the channel dimension can be better represented to meet the requirements of word generation. In the following three subsections, we first give the details of the design of pyramid feature maps to generate better attention regions, and a dual attention mechanism to re-calibrate the importance of the feature vector in the channel dimension by considering the hidden state of LSTM, then illustrate the overview of the proposed learning framework.

\subsection{Pyramid attention} 
\label{SEC:PA}

We introduce the pyramid attention module to make it capable of capturing multi-scale spatial properties in images. In a deep neural network, the size of receptive field roughly indicates how much information can be used to identify an object (e.g., {\em a car} or {\em a person}), some stuff (e.g., {\em grass} or {\em river}), or a composite visual patterns. The empirical receptive field in the convolutional output of a CNN is much smaller on high-level layers because the information is distilled, i.e., most of the visual properties are filtered in convolution operations. This makes the attention on a very small receptive field insufficiently incorporate the momentous prior to a receptive field in a larger size. So we address this issue by using the pyramid representation prior with multi-scale feature maps.

In a convolutional feature representation, one image is equally segmented into $w\times h$ grids, each of which describes a small region of the image. To produce the multiple vectorized features at different scales of the receptive field, we add extra average pooling modules to form the hierarchical feature maps by adopting varying-size pooling kernels. The coarsest level of average pooling generates the largest sub-regions of images to attend, and the following pyramid level separates the feature map into smaller areas. So the pyramid feature maps describe all possible sub-regions with varied sizes to attend for a better description. The number of pyramid levels and the size of each level in the encoder can be modified, which are related to the size of the feature map that is fed into the decoder. Different from the settings in \cite{CVPR17:PSPNET} that concatenates all pyramid feature maps in the channel dimension and the resolution is unchanged, we just increase the number of 2D regions at multi-scales in the encoder to generate rich feature maps. The multi-scale pooling kernel should maintain a reasonable gap in feature representation. In our setting, the pyramid attention is set to three levels with the bin sizes $1\times 1$ (original), $2\times 2$ and $4\times 4$, respectively. In the extreme case, global average pooling describes an image as a single visual feature vector, which is equivalent to the ``no attention'' model for image captioning \cite{CVPR15:IMCAP}. Extending the spatial attention to multi-scale pyramid attention can enrich the visual feature representation and sketch the hierarchical semantic pattern structures. Even if some synthesis patterns are completely useless for sentence generation, they can still be well suppressed by the RNN controller by setting to very low feature weights.

Recently, the bottom-up attention implemented by a pre-trained object detector can boost the caption performance due to the more accurate bounding boxes rather than the equal-sized grids \cite{CVPR18:BUTD}. For general image captioning purposes, such bottom-up attention with the help of auxiliary models can effectively improve the feature descriptive power. However, each visual feature vector in the bottom-up attention representation is no longer spatial-indicative. Even when the spatial properties collapse, the pyramid attention still applies. Note that the order of the local feature maps in the 2D feature representation can be arbitrarily changed, which does not affect the spatial attention, because the attention weights are solely dependent on the RNN hidden state and local feature representations. In this case, the pyramid attention becomes the visual feature synthesis that describes the composite visual patterns, which is similar to the scenario of the spatial-semantic search method proposed in \cite{CVPR17:VFS}. In the following parts of this article, we simply refer the attention on pyramid feature maps as pyramid attention ({\bf P-attention}).

\subsection{Dual attention}
\label{SEC:DA}

The spatial attention in Eq. (\ref{EQ:SA}) requires the visual feature map $\mathbf{V}$ and the hidden state $\mathbf{h}_{t-1}$ to calculate the weights in the width and height dimensions. However, the attention vector $\mathbf{v}_t^{(s)}$ is just a linear combination of $\mathbf{V}_i$ for $i=1,\ldots,L$, in which the channel dimension is unchanged. In the convolutional feature representation, each channel vector in a 2D grid or an object bounding box can be regarded as a word response, and different semantic responses are mutually associated with each other. By modelling the inter-dependencies among the 2D feature maps, it can improve the feature representation of specific semantics. Hence, to better represent the context vector to generate more accurate and smooth natural sentences, we enhance the feature map by introducing a dual attention module, which does not only focus on ``where to gaze'' in the spatial perspective, but also re-calibrate the importance of the feature components to improve the discriminative power.    

To discover the channel-wise dependency of the feature representation, we aim to learn a channel weight vector $\beta=[\beta_1,\ldots,\beta_d]\in\mathbb{R}^d$ to re-calibrate the feature map $\mathbf{V}$ thus form a channel context vector:
\begin{equation}
\mathbf{v}_t^{(c)} = \beta\odot \frac{1}{L}\sum\limits_{i=1}^L \mathbf{V}_i.
\end{equation}

Given a hidden state $\mathbf{h}_{t-1}$ of RNN at the time step $t-1$, the channel-wise score $\mathbf{c}$ and channel weight vector $\beta$ are computed in a similar way to the spatial attention as follows:
\begin{align}\label{EQ:DUAL_WEIGHT}
\mathbf{c} & = \tanh ((\mathbf{W}_c\mathbf{V}+\mathbf{b}_c)\oplus\mathbf{W}_{hc}\mathbf{h}_{t-1}), \nonumber \\
\beta & = \text{sigmoid}(\mathbf{W}_b \mathbf{c} + \mathbf{b}_b),
\end{align}
where $\mathbf{c}$ is the channel attention score vector. $\mathbf{W}_c$, $\mathbf{W}_{hc}$, $\mathbf{W}_b$ are learnable mapping matrices and $\mathbf{b}_c$, $\mathbf{b}_b$ are bias vectors, respectively. The main difference between Eq.(\ref{EQ:CE}) and Eq.(\ref{EQ:DUAL_WEIGHT}) is the linear projection terms $\mathbf{W}_s$ and $\mathbf{W}_c$, which project the feature map $\mathbf{V}$ from different dimensional perspectives. While the spatial attention vector $\alpha$ is to select which 2D regions can best describe the current word, the channel attention vector $\beta$ focuses on how much such features can generate the desired prediction, which can be considered as a channel regularization term. Also, in the channel attention, we use the sigmoid activation to re-calibrate the visual feature maps, but not the softmax function. This is mainly because each channel dimension works collaboratively with others, which is a joint feature representation mechanism but not ``one-hot'' vector-like to describe a small 2D region of images. Such a setting is similar to the self-attention used in \cite{AAAI:DESAN,CVPR18:SE}.

By applying the two attention mechanisms, the final context vector $\mathbf{v}_t$ at the time step $t$ to predict the word probability vector $\mathbf{y}_t$ is just the summation of the spatial context vector $\mathbf{v}^{(s)}_t$ and the channel context vector $\mathbf{v}^{(c)}_t$, i.e., 
\begin{equation}\label{EQ:DA}
\mathbf{v}_t=\mathbf{v}^{(s)}_t+\mathbf{v}^{(c)}_t.
\end{equation} 

When applying both spatial and channel attentions on the image feature representations, we name such an attention module as dual attention ({\bf D-attention}). 

Note that the dual attention scheme in our model differs from SCA-CNN \cite{CVPR17:SCA} from the following two perspectives: (1) The channel-wise attention and spatial attention in SCA-CNN are computed sequentially, while in our model, the two attentions from spatial and channel dimensions are computed in parallel. Inspired by the split-attention proposed in \cite{ARXIV:RESNEST}, a better visual feature representation can be obtained via an element-wise summation across multiple splits; (2) The activation of channel-wise attention used in SCA-CNN is a softmax function, which is the same with spatial attention in their paper. Such a setting is inappropriate because softmax is very like the one-hot encoding, which negatively affects the feature representation, so we applied the sigmoid function in the channel-wise attention. In the experimental part of this paper, we empirically prove that our dual attention can achieve much better results than SCA-CNN.


\begin{figure*}[t!]
\centering
    \includegraphics[width=0.9\textwidth]{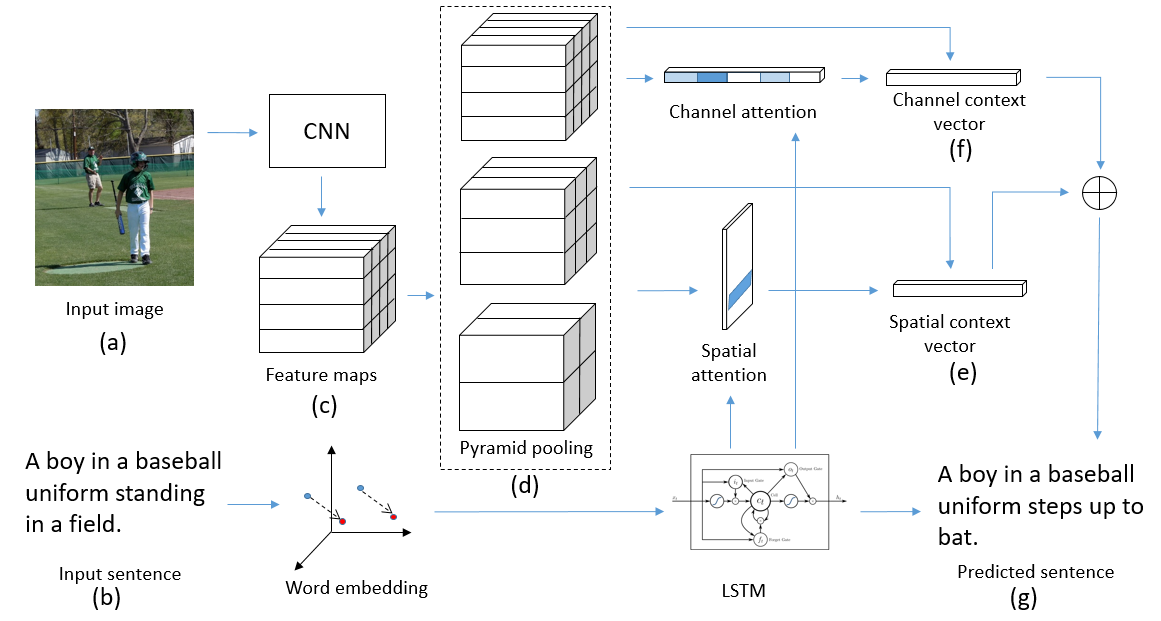}   
    \caption{The learning framework of the dual attention on pyramid feature maps for image captioning.}
\label{FIG:FRM}
\end{figure*}

\subsection{System overview}

Now we introduce how to equip pyramid attention and the dual attention module to a single image captioning model. The improvement the performance can benefit from the two modules, while limited computation resources overhead are required.

The attention model for image captioning is illustrated in Figure \ref{FIG:FRM}. Given an input image (a) and a sentence description (b), we use a pre-trained CNN (e.g., ResNet-101 \cite{CVPR16:RESNET}) or an object detector (e.g., the bottom-up visual features \cite{CVPR18:BUTD}) to extract the convolutional feature maps (c) and a word embedding layer for text representation. To better learn the word-sequence prediction, the visual feature maps are augmented by pyramid pooling (d). With the help of LSTM hidden state, the spatial and channel attentions are computed in parallel, forming a spatial context vector (e) and a channel context vector (f). By summing up the two vectors, they are fused as a final context vector for the word prediction in a partial sentence (g). 

The overall image captioning architecture in our work is based on the top-down attention model \cite{CVPR18:BUTD}, which contains two separate LSTMs: attention LSTM and language LSTM, respectively. In the sentence prediction, the input vector to the attention LSTM at each time step consists of the previous output of the language LSTM, concatenated with the global average pooling image feature $\overline{\mathbf{V}}=\frac{1}{L}\sum\limits_{i=1}^{L} \mathbf{V}_i$ and the encoding of the previous word:
\begin{equation}\label{EQ:INPUT_VEC}
\mathbf{x}_t^1=[\mathbf{h}_{t-1}^2, \overline{\mathbf{V}}, \mathbf{W}_e\amalg_t],
\end{equation}
where $\mathbf{W}_e$ is a word embedding matrix and $\amalg_t$ is the one-hot encoding of the input word at time step $t$. These inputs provide the attention LSTM with the maximal contextual information regarding the state of the language LSTM, the global feature of the image, and the partial sentence, respectively. Then the hidden state of the attention LSTM at the current time step $t$ is computed as:
\begin{equation}
\mathbf{h}_t^1 = \text{LSTM}(\mathbf{x}^1_t, \mathbf{h}^2_{t-1}).
\end{equation}

Given the hidden state $\mathbf{h}_t^1$, the spatial context vector $\mathbf{v}_t^{(s)}$ or the context vector of dual attentions $\mathbf{v}^t$ can be computed by Eq.(\ref{EQ:SA}) or Eq.(\ref{EQ:DA}), respectively (replacing $\mathbf{h}_t$ with $\mathbf{h}_t^1$ in Eq.(\ref{EQ:SPATIAL_WEIGHT}) and Eq.(\ref{EQ:DUAL_WEIGHT})). 

The input of the language LSTM is the concatenation of the context image feature and the output of the attention LSTM:
\begin{equation}
\mathbf{x}_t^2 = [\mathbf{v}_t, \mathbf{h}^1_t],
\end{equation}
thus the hidden state output of the language LSTM becomes:
\begin{equation}
\mathbf{h}_t^2 = \text{LSTM}(\mathbf{x}^2_t, \mathbf{h}^1_t).
\end{equation}

Using the notation $Y$ in Eq.(\ref{EQ:Y}), at each time step $t$ the conditional distribution over possible words is computed by:
 
\begin{equation}\label{EQ:PREDICTION}
p(\mathbf{y}_t|\mathbf{h}^2_{t}) = \text{softmax}(\mathbf{W}_p\mathbf{h}^2_t+\mathbf{b}_p),
\end{equation}
where $\mathbf{W}_p$ and $\mathbf{b}_p$ are trainable weights and biases, respectively. The word distribution over complete sentence is calculated as the product of conditional distributions:
\begin{equation}\label{EQ:PROB}
p(Y)=\prod\limits_{t=1}^{T}p(\mathbf{y}_t|\mathbf{h}^2_{t}).
\end{equation}

Assume the whole trainable parameter set is $\theta$. When given a reference sentence represented by $\hat{\mathbf{y}}_{1:T}$, the most straight-forward way to optimize the captioning model is to minimize the cross-entropy loss of each individual word:
\begin{equation}\label{EQ:CE}
L_{CE}(\theta)=-\sum\limits_{t=1}^{T} \log(p_{\theta}(\mathbf{y}_t^*|\mathbf{y}_{1:t-1}^*)).
\end{equation}

Though training with the cross-entropy loss enables the fully diffirentiable optimization by backpropagation, the training objective is inconsistent with the language evaluation metrics (e.g., CIDEr score). Furthermore, this creates a schism between training and evaluation because in the inference the model has no access to the previous ground truth token, which leads to cascading errors and biased semantics. So after the cross-entropy optimization, we can use the reinforcement learning to minimize the negative expected reward:

\begin{equation}
L_{CD}(\theta)=-\mathbb{E}_{\mathbf{y}_{1:T}\sim p_{\theta}}[r(\mathbf{y}_{1:T})],
\end{equation}
where $r(\cdot)$ is the reward function. It has been proved that using the CIDEr score in Self-Critical Sequence Training (SCST) \cite{CVPR17:SCSIC} can effectively generate sentences with higher quality. The approximate gradient is computed as follows:
\begin{equation}
\nabla_{\theta}L_{CD}(\theta)\approx - (r(\mathbf{y}^s_{1:T})-r(\hat{\mathbf{y}}_{1:T}))\nabla_{\theta}\log(r(\mathbf{y}^s_{1:T})),
\end{equation}
where $\mathbf{y}_{1:T}$ is a sampled caption and $\hat{\mathbf{y}}_{1:T}$ is the baseline score calculated by greedy decoding. Fine-tuning the captioning model with reinforcement learning focuses each prediction step to achieve the best score of the overall sentence and explores the space of captions by sampling from the policy, whose gradient tends to increase the probability of sampled captions with a higher CIDEr score. Thus, after a few epochs' fine-tuning, the CIDEr score can be significantly increased, which also benefits all other evaluation metrics.

In the inference procedure, given an image $I$, the sentence is generated in the word-by-word manner. The variables $\mathbf{h}^1_0$ and $\amalg_0$ are initialized by a zero vector and the one-hot encoding of the start token, respectively. Each word $\mathbf{y}_t$ is sequentially calculated by applying Eq.(\ref{EQ:INPUT_VEC}) - Eq.(\ref{EQ:PREDICTION}) until the end token of the sentence is met. However, alternatively using Eq.(\ref{EQ:PREDICTION}) as a greedy search may be a sub-optimal choice in Eq.(\ref{EQ:PROB}). Usually, we use beam search, which selects multiple alternatives for an input sequence at each timestep based on conditional probability. The beam search is also used for sampling $\mathbf{y}_{1:T}$ in the self-critical fine-tuning.

\section{Experiments}
\label{SEC:EXP}

We describe the experimental settings and results, then qualitatively and quantitatively validate the effectiveness of the proposed dual attention on pyramid feature maps method for image captioning. We validate the effectiveness of the proposed dual attention on pyramid feature maps method for image captioning, by answering the following two questions: {\bf Q1:} Are pyramid attention and dual attention effective to prepare better visual feature representations for content description in images? {\bf Q2:} How does the performance of our method compare to other state-of-the-art image captioning methods in a single model?

\subsection{Datasets}

We report the experimental results on three widely-used benchmarks: (1) Flickr8K \cite{JAIR:FLICKR8K}: this dataset is comprised of 8000 images in total and is split into 6,000, 1,000, and 1,000 images for training, validation and testing, respectively; (2) Flickr30K \cite{ACL14:FLICKR30K}: this is a larger dataset with 25,381, 3,000, 3,000 for training, validation and testing, respectively. In both of the above two Flickr datasets, each image is manually annotated by 5 sentences, and we report the results according to the standard splits. (3) MS COCO \cite{ECCV14:COCO}: it is the largest image captioning dataset as far as we know, which contains 164,042 images in total. As the ground truth of the test set is withheld by the organizers, we followed \cite{CVPR15:KARPATHY} to use the 82,783 training set to learn the model, 5,000 images for validation and another 5,000 images for testing, respectively.

\subsection{Implementation details and computational complexities}

We used three types of convolution features as image representations: the final convolutional output of VGG19 \cite{ICLR15:VGG} ResNet-101, ResNet-152 \cite{CVPR16:RESNET} on the Flickr8K and Flickr30K datasets, respectively. When the input RGB image has the resolution $224\times224$, the image is equally segmented into 49 grids, each of which describes a specific region of the image. On the MS COCO dataset, we also used the bottom-up feature computed by a pre-trained object detector \cite{CVPR18:BUTD}. To produce the multiple vectorized feature representation at different scales of the receptive field, we added a $2\times2$ and a $4\times4$ average pooling, both of which were conducted on the original convolution output with 1 stride but no padding. Thus, the extra two pooling operations give a $6\times6\times d$ and a $4\times4\times d$ feature map, respectively. Here $d=512$ for VGG19 and $d=2048$ for ResNet. On the pyramid representation of convolutional features, we applied separate dense mappings with ReLU activations to reduce the feature dimensionality to 512. On the MS COCO dataset, we also used the bottom-up features provided by \cite{CVPR18:BUTD}, with the fixed $36\times2048$ feature map size. For language processing, we used a word embedding layer to map each word in a description sentence to a 512d vector. In both of the top-town attention and language LSTMs, the sizes of the hidden units were set to 512 without any change.


To observe the performance gains of the two proposed attention methods, we tested two soft-attention-based image captioning models, Show-Attend-Tell \cite{ICML15:SAT} and Top-down \cite{CVPR18:BUTD}, on the Flickr8K and Flickr30K datasets. Specifically, we conducted the experiments using P-attention and D-attention separately. After that, we combined them as a single model (P+D attention) to test its effectiveness on improving the quality of generated sentences. All of the three attention models were trained with stochastic gradient descent using the AdamW optimizer \cite{ICLR19:ADAMW}. We followed \cite{CVPR18:DODC} to add a discriminability loss in training image captioning models to improve the quality of resulting sentences. The batch size was set to 96, which can well fit the memory of a single Titan Xp GPU card. On Flickr8K and Flickr30K datasets, we only conducted the cross-entropy optimization with convolutional features to test the effectiveness of the proposed methods. On the MS COCO dataset, we set a two-stage training procedue. At the first stage, we used the BLEU-4 score in the validation for model selection in minimizing cross-entropy, which finished within 60 epochs. At the second stage, we fine-tuned the captioning model by optimizing Eq. (\ref{EQ:CE}) with 30 epochs. In the inference procedure, we used the beam search (with beam size 5) to generate the best natural sentences.

The feature maps of images were all pre-computed and cached to accelerate the training procedure. Assume we use the feature map of ResNet-101 with size $7\times7\times2048$, the sentence length is fixed to 50, and the vocabulary size is 8,000, the numbers of trainable parameters and FLOPs are summarized in Table I. Imposing either P-attention or D-attention only leads to a marginal increase of learnable parameters. In terms of FLOPs, the computational resources overhead of P-attention mainly incurred by multiple linear mapping functions on the convolution feature maps, while D-attention requires more resources in computing the channel attention, where the dimension is much higher than the number of 2D visual regions. Actually, there is always a trade-off between model performance and computational complexities. Applying the two attentions separately or building a unified model ({\bf P+D attention}), although has a higher computational complexity in the training process, it can still well fit most GPUs, with the improvement of the better quality in image-language translation.

\begin{table}[h]
\centering \small
\caption{Computational complexities.}
\label{TB:FLOP}
\begin{tabular}{|c||c|c|}
\hline
Model		& Params & FLOPs \\
\hline
Top-down attention     	& 29.5M    & 1.16G \\
Top-down + P-attention	    	& 29.6M    & 1.99G  \\
Top-down + D-attention	      & 31.8M    & 2.68G\\
Top-down + P+D attention	 & 31.9M    & 5.32G \\
\hline
\end{tabular}
\end{table}

\subsection{Evaluation metrics}  

We use BLEU scores ({\bf B@1}, {\bf B@2}, {\bf B@3} and {\bf B@4}) \cite{ACL02:BLEU} without brevity penalty to evaluate the image caption generation. Due to the criticism of BLEU, we also report the METEOR score ({\bf MT}) \cite{ACL05:METEOR}. All of the above metrics were evaluated by the NLTK toolkit\footnote{\url{https://www.nltk.org}}. For the evaluation on MS COCO dataset, we also use ROUGE-L ({\bf RG-L}) \cite{ACL04:ROUGE} and CIDEr \cite{CVPR15:CIDER} for evaluation\footnote{Note that the computations of BLEU and METEOR with the nlg-eval package (\url{https://github.com/Maluuba/nlg-eval}) differs from NLTK.}.

\subsection{Qualitative results}

\begin{figure*}[t]
\centering
\begin{subfigure}{0.68\textwidth}
\includegraphics[width=\textwidth]{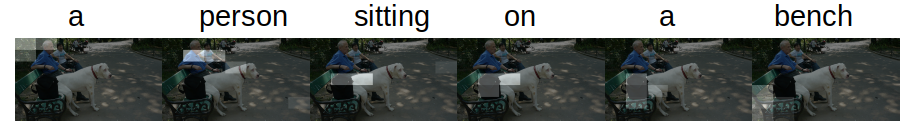}        
\caption{Single-scale attention with ResNet-101.}
\end{subfigure}
\begin{subfigure}{0.83\textwidth}
\includegraphics[width=\textwidth]{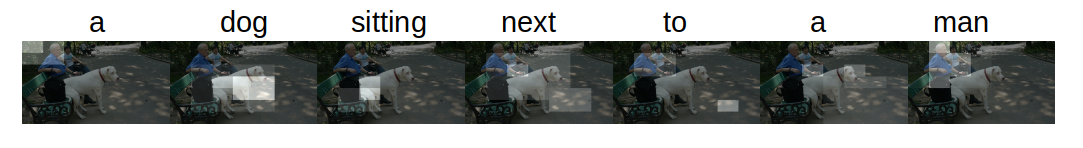}        
\caption{Pyramid attention with ResNet-101.}
\end{subfigure}
\begin{subfigure}{1.0\textwidth}
\includegraphics[width=\textwidth]{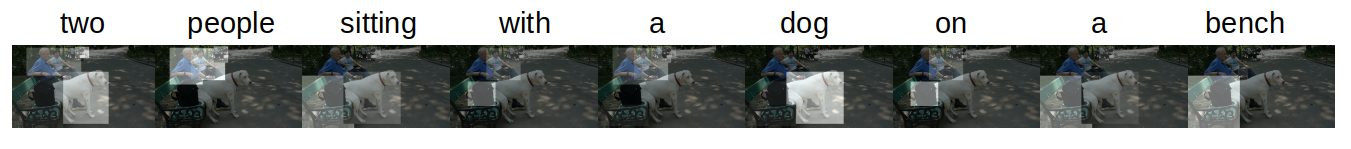}        
\caption{Pyramid attention with bottom-up features.}
\end{subfigure}    
\caption{ A visualization of top-down attentions on salient areas.}
\label{FIG:PYRAMID_ATT}
\end{figure*}
\begin{figure*}[t]
\centering
\includegraphics[width=0.8\textwidth]{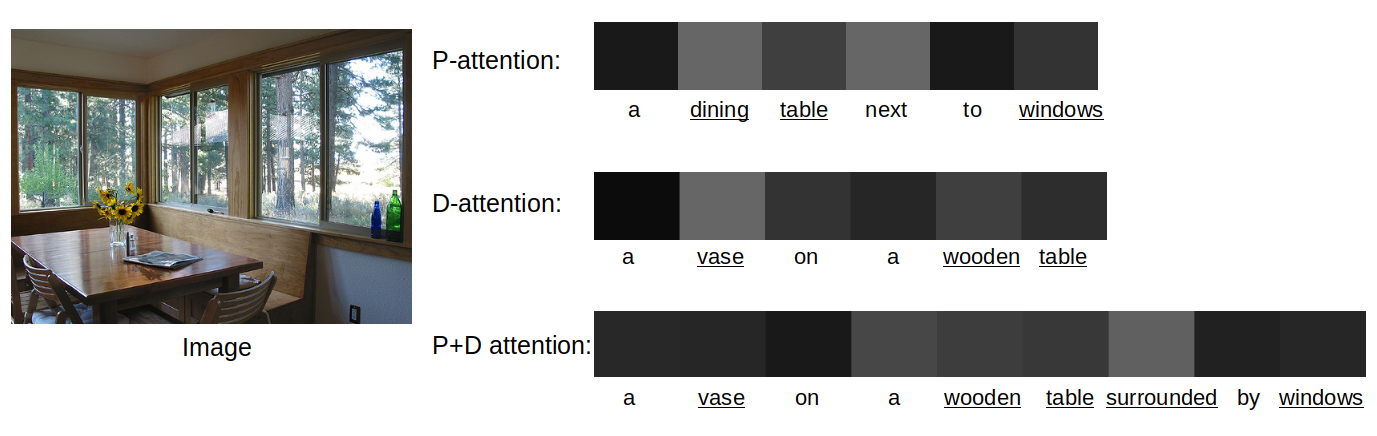} 
\caption{A visualization of word predictions (ResNet-101).}
\label{FIG:DUAL_ATT}
\end{figure*}

We show the P-attention and D-attention results as in Figure \ref{FIG:PYRAMID_ATT} and \ref{FIG:DUAL_ATT}, respectively. By using the convolution features, the saliency regions can show more accurate attentions on the original image, since the pyramid attention provides a broader range of feature maps to attend. Using the bottom-up feature with pyramid attentions can not only describe the visual content {\em people} and {\em dog}, but also the details of the contextual clue {\em bench}. By applying the D-attention, we observe the improvement on the model discriminative ability. For example, the higher confidence scores on some nouns such as {\em windows} and {\em wooden table}.

In Table \ref{TB:CAP_EXAMP}, we give some captioning result examples of the proposed attention model. In the comparison of the generated sentences by the proposed method and the ground truth sentences, all our three attention methods can well describe the most important spatial and semantic information of the images to meet the humans' cognition. The pyramid attention can generally sketch the hierarchical visual properties and identify the important entities, while the dual attention can give more detailed and accurate descriptions of the entities. The two types of attention are complementary to each other, so the joint learning framework generally outperforms any of the two single attention models.

\begin{table*}[t!]
\centering \small
\caption{Examples of image captioning using the proposed methods (using ResNet-101).}
\label{TB:CAP_EXAMP}
\begin{tabular}{|L{4cm}|L{4cm}|L{4cm}|L{4cm}|}
\hline
    \begin{minipage}{0.22\textwidth}
    \includegraphics[width=1\textwidth, height=0.65\textwidth]{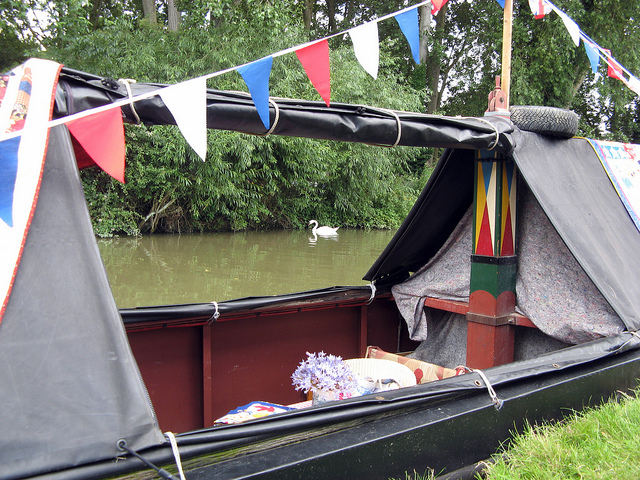}
    \end{minipage}
&
    \begin{minipage}{0.22\textwidth}
    \includegraphics[width=1\textwidth, height=0.65\textwidth]{figs/COCO_val2014_000000418281.jpg}
    \end{minipage}
&
    \begin{minipage}{0.22\textwidth}
    \includegraphics[width=1\textwidth, height=0.65\textwidth]{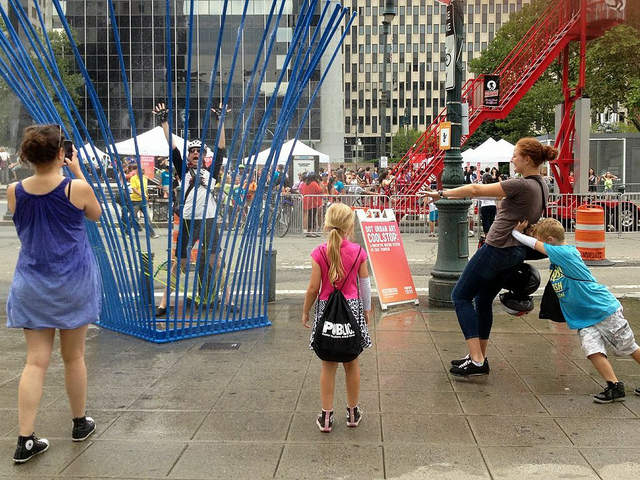}
    \end{minipage}
&
    \begin{minipage}{0.22\textwidth}
    \includegraphics[width=1\textwidth, height=0.65\textwidth]{figs/COCO_val2014_000000483108.jpg}
    \end{minipage}
\\
{\bf P-attention:} a boat and a swan parking on a river. &{\bf P-attention:} two cows sitting on the grass with a building. &{\bf P-attention:} a group of people playing in the courtyard. &{\bf P-attention:} a person riding a bicycle with a train. \\
{\bf D-attention:} a boat with flags sitting next to a green shore. &{\bf D-attention:} a cow is standing on a grassland. &{\bf D-attention:} a girl with a red shirt standing on the ground  &{\bf D-attention:} a person on a bicycle next to a train.\\
{\bf P+D attention:}  a swan and a boat floating on the river. &{\bf P+D attention:} two cows standing on the grass with a temple. &{\bf P+D attention:} a group of people playing in the city street. &{\bf P+D attention:} a man is riding a bike in front of a red and white train.\\
{\bf GT:}  a swan is floating down the river by the boat.  &{\bf GT:} two cows outside one laying down and the other standing near a building. &{\bf GT:} people are standing outside in a busy city street. &{\bf GT:} a red and white train and a man riding a bicycle.\\
\hline
\end{tabular}
\end{table*}

\subsection{Quantitative results}

\begin{table}[h]
\centering \small
\caption{METEOR scores with pyramid poolings.}
\label{TB:PYRAMID_POOLINGS}
\begin{tabular}{|c||c|c|}
\hline
Pooling	settings & Flickr8K & Flickr30K \\
\hline
 $1\times 1$ ($L=49$) 	& 22.8  & 27.7 \\
 $2\times 2$ ($L=36$)	& 20.4  & 25.5 \\
 $4\times 4$ ($L=16$)	& 18.9   & 24.6 \\
All & 23.5 & 28.1 \\
\hline
\end{tabular}
\end{table}

\begin{table}[h]
\centering \small
\caption{METEOR scores with dual attentions.}
\label{TB:CS_DUAL}
\begin{tabular}{|c||c|c|}
\hline
Attention	settings & Flickr8K & Flickr30K \\
\hline
SC 	& 29.5 & 27.6 \\
CS	& 29.7 & 27.5 \\
Parallel (ours) & 30.2 & 28.2 \\
\hline
\end{tabular}
\end{table}

\begin{table*}[h]
\centering \small
\caption{Performance comparison with the state-of-the-arts on the Flickr8K and Flickr30K datasets. All image captioning models are trained using convolution features.}
\label{TB:COMP_FLICKR}
\begin{tabular}{|c|c|c|c|c|c|c|c|}
\hline
{\bf Dataset} & {\bf Method} & {\bf Feature}	& {\bf B@1}	& {\bf B@2}	& {\bf B@3}	& {\bf B@4}& {\bf MT} \\
\hline
\multirow{16}{*}{Flickr8K} &Show-Att-Tell\cite{ICML15:SAT} & VGGNet	 & 67.0 & 44.8 & 29.9 & 19.5 & 18.9   \\
&SCA-CNN\cite{CVPR17:SCA}	& ResNet-152 & 68.2 & 49.6 & 35.9 & 25.8  & 22.4  \\ 
&Bi-LSTM\cite{MM16:BLSTM}	& VGG-16 & 65.5 & 46.8 & 32.0 & 21.5 & -  \\
\cline{2-8}
&Show-Att-Tell (Our implementation) &ResNet-101	& 67.9 & 48.6 & 33.7  & 20.9  & 21.4  \\ 
&Show-Att-Tell + P-attention (Ours) &ResNet-101	& 68.1 & 49.5 & 34.2  & 21.3  & 21.5  \\ 
&Show-Att-Tell + D-attention (Ours) &ResNet-101	& 68.5 & 49.9  & 36.8  &  22.6 & 22.5   \\ 
&Show-Att-Tell + P+D attention (Ours) &ResNet-101	& 68.4 & 50.3  & 37.0  & 22.8  & 22.6  \\ 
&Top-down + P-attention (Ours) &VGG-19	& 67.4 & 52.1  & 44.8  & 36.6   & 23.5  \\ 
&Top-down + D-attention (Ours) &VGG-19	& 67.7 &  53.2 & 45.1 & 36.8  &   23.9\\ 
&Top-down + P+D attention (Ours) &VGG-19	& 67.8 & 53.3  & 45.2  & 37.0  & 24.2  \\ 
&Top-down + P-attention (Ours)  &ResNet-101	& 69.1 & 57.5 & 47.5 & 39.5 & 29.4   \\
&Top-down + D-attention (Ours) &ResNet-101 & 69.8 & 58.5 & 48.7 & 40.7 & 30.2   \\
&Top-down + P+D attention (Ours) &ResNet-101 & {\bf 70.0} & 59.2 & {\bf 49.4} & 41.4 & {\bf 30.9}  \\
&Top-down + P-attention (Ours) &ResNet-152	& 69.2  & 57.7  & 47.2  & 39.4  & 29.3  \\ 
&Top-down + D-attention (Ours) &ResNet-152	& 69.7 & 59.0  & 48.8  & 40.7  & 30.6 \\ 
&Top-down + P+D attention (Ours) &ResNet-152 & 69.9 & {\bf 59.5} & 49.3 & {\bf 41.6}  & 30.7  \\ 
\hline
\multirow{23}{*}{Flickr30K} & Show-Att-Tell\cite{ICML15:SAT} & VGGNet	& 66.7 & 43.4 & 28.8 & 19.1 & 18.5  \\
& ATT-FCN\cite{CVPR16:SA} & GoogleNet & 64.7 & 46.0 & 32.4 & 23.0 & 18.9 \\
& SCA-CNN\cite{CVPR17:SCA}	& ResNet-152 & 66.2 & 46.8 & 32.5 & 22.3 & 19.5  \\
& Bi-LSTM\cite{MM16:BLSTM}	& VGG-16 & 62.1 & 42.6 & 28.1 & 19.3 & - \\
& Saliency+Context Attention\cite{TOMM:AS} & VGGNet & 61.5 & 43.8 & 30.5 & 21.3 & 20.0 \\
& Attention Correctness\cite{AAAI17:AC} & VGGNet & - & - & 38.0 & 28.1 & 23.0 \\
& Language CNN\cite{ICCV17:LCNN} & VGGNet & 73.8 & 56.3 & 41.9 & 30.7 & 22.6 \\	& Adaptive attention\cite{CVPR17:ADA} & ResNet	 & 67.7 & 49.4 & 35.4 & 25.1 & 20.4 \\
& Att2in+RD\cite{TMM:STP} &	Resnet-101 & - & - & - & - & 26.0 \\
& hLSTMat\cite{TPAMI:HLSTMAT} &	ResNet-101 & 73.8 & 55.1 & 40.3 & 29.4 & 23.0 \\
\cline{2-8}
&Show-Att-Tell (Our implementation) &ResNet-101	& 69.4  & 45.6  & 32.9 & 22.4  & 19.6  \\ 
&Show-Att-Tell + P-attention (Ours) &ResNet-101	& 69.6  & 45.6  & 34.1  & 22.8   & 19.9   \\ 
&Show-Att-Tell + D-attention (Ours) &ResNet-101	& 69.9 & 46.1  & 35.3  & 23.5   & 20.2  \\ 
&Show-Att-Tell + P+D attention (Ours) &ResNet-101	& 69.7 & 46.1  & 35.5 & 23.7  & 20.4  \\ 
&Top-down + P-attention (Ours) &VGG-19 & 69.2  & 57.7 & 50.2  & 26.5  & 24.1  \\ 
&Top-down + D-attention (Ours) &VGG-19 & 69.8 & 58.5 & 52.1 & 27.4  & 24.5  \\ 
&Top-down + P+D attention (Ours) &VGG-19 & 71.3 & 58.6 & 52.2 & 33.9  & 25.6  \\ 
&Top-down + P-attention (Ours)  &ResNet-101	& 72.3 & 62.7 & 53.3 & 26.0 & 28.1 \\
&Top-down + D-attention (Ours)  &ResNet-101	& {\bf 76.5} & 64.2 & 53.3 & 44.6 & 28.2 \\
&Top-down + P+D attention (Ours)  &ResNet-101	& 74.4 & 64.8 & {\bf 55.5} & 47.8 & 28.7  \\
&Top-down + P-attention (Ours) &ResNet-152	& 73.5 & 62.8 & 52.7  & 43.9   & 28.2  \\ 
&Top-down + D-attention (Ours) &ResNet-152	& 76.3  & 64.7  & 55.4  & 46.6  & {\bf 28.8}  \\ 
&Top-down + P+D attention (Ours) &ResNet-152	& 76.1 & {\bf 64.9} & 55.3 & {\bf 47.8}  & 28.3  \\ 
\hline
\end{tabular}
\end{table*}

We first give the ablation study on Flickr8K and Flickr30K to investigate how the proposed two attentions provide better feature representation priors for image captioning. For pyramid attention, we adopted three pooling settings, $1\times 1$, $2\times 2$ and $4\times 4$ separately with P-attention model only, then combined them together as a pyramid pooling setting. Note that only using $1\times 1$ is equivalent to the Bahdanau attention. We recorded the METEOR scores, as are shown in Table \ref{TB:PYRAMID_POOLINGS}. When increasing the pooling scale, the performance essentially degrades. This is reasonable because the larger receptive fields often contain fewer local visual details to predict the semantics. We then conducted the ablation study of the dual attention. Following \cite{CVPR17:SCA}, we tested the sequential attentions of spatial-channel ({\bf SC}) and channel-spatial ({\bf CS}), as well as the proposed parallel structure. The METEOR scores summarized in Table \ref{TB:CS_DUAL} empirically prove that the split attention can achieve better performance. 

We summarize the statistical results of the proposed P-attention, D-attention and P+D attention methods, then compare with state-of-the-art image captioning models in Table \ref{TB:COMP_FLICKR} and Table \ref{TB:COMP_COCO} for the Flickr8K, Flickr30K and MS COCO datasets, respectively. In the two tables, we use the bold font to emphasize the best results.

By observing the results in Table \ref{TB:COMP_FLICKR} obtained by two separate attentions introduced in Section \ref{SEC:PA} and \ref{SEC:DA}, we can get the following findings: 
\begin{itemize}
\item The visual feature plays a significant role in the image captioning performance. The ResNet features are generally more descriptive in translating image contents to natural sentences.
\item Similar to other supervised learning tasks, the baseline model plays an important role in the performance of image captioning. Using both attention and language LSTM in the top-down model can generate better captioning results compared to Show-Attend-Tell.
\item Applying the top-down model using both attention and language RNN modules can achieve better captioning performance compared to a single RNN in the decoder.
\item Both pyramid attention and dual attention can provide better feature representation prior, compared to the Bahdanau attention for visual content captioning. 
\end{itemize}
Specifically, on the Flickr8K dataset, the joint attention model (P+D attention) outperforms the second-best method by 1.8 / 9.9 / 13.5 / 15.8 / 8.5 percent in terms of BLEU-1, BLEU-2, BLEU-3, BLEU-4 and METEOR, respectively. On the larger Flickr30K dataset, the joint attention model (P+D attention) outperforms the second-best method by 2.7 / 8.6 / 13.6 / 17.1 / 2.8 percent in terms of the five evaluation metrics, respectively. For the results on this dataset, simply applying the dual attention obtains very high scores in terms of BLEU and METEOR. The composite model with P+D attention can generally perform better results compared to the two single attentions.

On the large-scale MS COCO dataset, we separately report the model performance comparisons using convolution features and bottom-up features. Also, we reported the evaluation results of both cross-entropy optimization and CIDEr fine-tuning. When using convolution features, both pyramid attention and dual attention can effectively improve the baseline (Top-down). After fine-tuning, the performance is further boosted.

The very recent image captioning models \cite{CVPR18:BUTD,CVPR19:LBPF,ICCV19:HIP,ICCV19:AOA} adopt the bottom-up attention instead of the traditional convolutional features to accurately implement the visual-semantic mapping. With the help of a pre-trained object detector to better attend the visual feature maps within the bounding boxes, these models achieve a noteworthy improvement over the captioning models that using convolutional feature maps. However, it requires that the training sources for both detection and description are homogeneous. In some specific cases when the two data sources are heterogeneous, e.g., the dataset for the object detector training contains no person, while the target captioning dataset is mainly about human action description, the bottom-up attention will fail.

In our experiment, the pyramid attention on bottom-up features can be considered as multi-scale semantic aggregation, which can slightly improve the performance compared to the up-down method proposed in \cite{CVPR18:BUTD}. The dual attention, however, is more effective to prepare better visual feature representation for image captioning. Our composite attention method, i.e., P+D attention model trained with bottom-up features, achieved very competitive results. Note that our methods P-attention, D-attention and P+D attention are highly modular and can be integrated into a variety of image captioning frameworks. In this work, our models are based on the up-down model, while all other comparison methods improve the attention-based image captioning models from different perspectives. Thus, the performance of the proposed image captioning model can be further improved if we use more advanced settings. For example, using the ``look back and predict forward'' strategy \cite{CVPR19:LBPF}. In our experiments, we have mainly proved that the proposed two attention variants P-attention and D-attention, as well as the joint learning approach P+D attention, are able to augment the computational capacity by fully exploring the visual-semantic relationships between visual image features and natural languages.

\begin{table*}[t!]
\centering \small
\caption{Performance comparison with the state-of-the-art on MS COCO dataset. All mothods are evaluated in the single-model mode. The methods with \textdagger are trained with the bottom-up attention, i.e., the visual features are obtained by an auxiliary object detection model.}
\label{TB:COMP_COCO}
\begin{tabular}{|c|c|c|c|c|c|c|c|c|c|c|c|}
\hline
\multirow{2}{*}{{\bf Method}} & \multicolumn{5}{c|}{Cross-entropy optimization} & \multicolumn{5}{c|}{CIDEr fine-tuning} \\
\cline{2-11}
& {\bf B@1} &{\bf B@4}& {\bf MT} & {\bf RG-L} & {\bf CIDEr} &{\bf B@1}	&{\bf B@4}& {\bf MT} & {\bf RG-L} & {\bf CIDEr}\\
\hline
SCA-CNN\cite{CVPR17:SCA}		& 71.9 & 31.1 & 25.0 & - & - & - & - & - & - & - \\
Adaptive attention\cite{CVPR17:ADA} & 74.2 & 33.2 & {\bf 26.6} & - & {\bf 108.5} &- &- &- &- &- \\
Semantic guidance\cite{TMM:SEMANTIC} & 71.2 & 26.5 & 24.7 & - & 88.2 & - & - & - & - & - \\
Att2in+RD\cite{TMM:STP} & - & 34.3 & 26.4 & {\bf 55.2} & 106.1 & - & 35.2 & 27.0 & 56.7 & {\bf 115.8} \\ 
SCST\cite{CVPR17:SCSIC} 	&- &- &- &- &- & - & 31.3 & 26.0 & 54.3 & 101.3 \\
Top-down\cite{CVPR18:BUTD}  & 74.5 & 33.4 & 26.1 & 54.4 & 105.4 &76.6 &34.0 &26.5 &54.9 &111.1 \\
P-attention (Ours)  	& 75.1 & 33.8  & 26.0 & 54.2 & 105.1 &77.7 &35.6 &27.2 &56.5 &113.3\\
D-attention (Ours)  	& 75.7 & 34.7  & 25.8 & 54.7 & 106.7 & {\bf 78.6} &35.9 &{\bf 27.7} &57.6 &114.0 \\
P+D attention (Ours)  & {\bf 76.4} & {\bf 35.5} & 26.2 & 54.8 & 108.4 &78.3 &{\bf 36.5} &27.3 & {\bf 57.8} & 114.4 \\

\hline

Up-down\cite{CVPR18:BUTD}\textdagger  & 77.2 & 36.2 & 27.0 & 56.4 & 113.5 & 79.8 & 36.3 & 27.7 & 56.9 & 120.1  \\
Object relationship\cite{TMM:IMCAP_SG}\textdagger & 76.7 & 33.8 & 26.2 & 54.9 & 96.5 & 79.2 & 36.3 & 27.6 & 56.8 & 120.2 \\ 
hLSTMat\cite{TPAMI:HLSTMAT}\textdagger & - & - & - & - & - & - & 37.5 & 28.5 & 58.2 & 125.6  \\
GCN-LSTM\cite{ECCV18:GCN_LSTM}\textdagger & 77.4 & 37.1 & 28.1 & 57.2 & 117.1 & {\bf 80.9} & 38.3 & 28.6 & 58.5 & 128.7  \\
Attenion on Attention\cite{ICCV19:AOA}\textdagger	& 77.4 & 37.2 & 28.4 & 57.5 & {\bf 119.8} & 80.2 & 38.9 & 29.2 & 58.8  & 129.8 \\
Up-down + HIP\cite{ICCV19:HIP}\textdagger	& - & 37.0 & 28.1 & 57.1 & 116.6 & - & 39.1 & 28.9 & 59.2 & {\bf 130.6} \\
Up-down + RD\cite{TMM:STP}\textdagger & - & 36.7 & 27.8 & 56.8 & 114.5 & - & 37.8 & 28.2 & 57.9 & 125.3 \\
Adaptive attention time\cite{NIPS19:AAT}\textdagger	& - & 37.0 & 28.1 & 57.3 & 117.2 & - & 38.7 & 28.6 & 58.5 & 128.6 \\
Global-Local Discriminative Objective \cite{TMM:GLD} \textdagger &- &- &- &- &- & 78.8 & 36.1 & 27.8 & 57.1 & 121.1 \\
LBPF\cite{CVPR19:LBPF}\textdagger	& {\bf 77.8} & {\bf 37.4} & 28.1 & 57.5 & 116.4 & 80.5 & 38.3 & 28.5 & 58.4 & 127.6 \\
P-attention (Ours)\textdagger  	& 77.1  & 36.5  & 27.4  & 56.6  & 114.3 & 79.3 & 38.5 & 28.8  & 58.1 & 124.0 \\
D-attention (Ours)\textdagger  	& 77.3  & 36.9  & 28.5 & 57.2  & 116.4 & 80.7 & {\bf 39.8} & 29.4 & 58.8 & 129.8  \\
P+D attention (Ours)\textdagger  & {\bf 77.8} & 37.3  & {\bf 28.6}  & {\bf 57.6}  & 117.3 & 79.8 & 39.4 & {\bf 30.1} & {\bf 59.4} & 129.5 \\
\hline
\end{tabular}
\end{table*}



\section{Conclusion}
\label{SEC:CONCLUSION}

We have presented a novel learning framework that applies dual attention on pyramid feature maps for image captioning. Different from the attention model \cite{ICCV19:AOA} mainly focuses on language generation, we try to explore visual feature representations for image captioning. The P-attention model takes attentions from multi-scale receptive fields in an image, which is a reasonable way to improve the visual feature representation by enriching the feature maps. The proposed D-attention model, on the other hand, can better leverage the channel-wise and spatial-wise features from two different perspectives. Both P-attention and D-attention models can boost the image captioning performance, while jointly applying the two proposed modules to form a unified learning framework, our P+D attention model achieves the state-of-the-art performance in a single captioning model. We believe that our work can also benefit the research community of visual-semantic understanding in other learning tasks such as visual question and answering (VQA) \cite{MM18:VQA} and video captioning \cite{MM17:MMA}, and we will also explore the potential of the proposed method in both network structure and other application fields. 

\bibliographystyle{IEEEtran}
\bibliography{sample-base}

\begin{thebibliography}{10}
\providecommand{\url}[1]{#1}
\csname url@samestyle\endcsname
\providecommand{\newblock}{\relax}
\providecommand{\bibinfo}[2]{#2}
\providecommand{\BIBentrySTDinterwordspacing}{\spaceskip=0pt\relax}
\providecommand{\BIBentryALTinterwordstretchfactor}{4}
\providecommand{\BIBentryALTinterwordspacing}{\spaceskip=\fontdimen2\font plus
\BIBentryALTinterwordstretchfactor\fontdimen3\font minus
  \fontdimen4\font\relax}
\providecommand{\BIBforeignlanguage}[2]{{%
\expandafter\ifx\csname l@#1\endcsname\relax
\typeout{** WARNING: IEEEtran.bst: No hyphenation pattern has been}%
\typeout{** loaded for the language `#1'. Using the pattern for}%
\typeout{** the default language instead.}%
\else
\language=\csname l@#1\endcsname
\fi
#2}}
\providecommand{\BIBdecl}{\relax}
\BIBdecl

\bibitem{MM17:ACMR}
B.~Wang, Y.~Yang, X.~Xu, A.~Hanjalic, and H.~T. Shen, ``Adversarial cross-modal
  retrieval,'' in \emph{MM}, 2017, pp. 154--162.

\bibitem{TIP:LDB}
X.~Xu, F.~Shen, Y.~Yang, H.~T. Shen, and X.~Li, ``Learning discriminative
  binary codes for large-scale cross-modal retrieval,'' \emph{IEEE Trans. on
  Image Processing}, vol.~26, no.~5, pp. 2494--2507, 2017.

\bibitem{TPAMI:LSRH}
K.~Li, G.-J. Qi, J.~Ye, and K.~A. Hua, ``Linear subspace ranking hashing for
  cross-modal retrieval,'' \emph{IEEE trans. on pattern analysis and machine
  intelligence}, vol.~39, no.~9, pp. 1825--1838, 2016.

\bibitem{TPAMI:DA}
B.~Zhang, D.~Xiong, and J.~Su, ``Neural machine translation with deep
  attention,'' \emph{IEEE trans. on pattern analysis and machine intelligence},
  2018.

\bibitem{JTNMT:ABJT}
Y.~Cheng, ``Agreement-based joint training for bidirectional attention-based
  neural machine translation,'' in \emph{Joint Training for Neural Machine
  Translation}, 2019, pp. 11--23.

\bibitem{ICML15:SAT}
K.~Xu, J.~Ba, R.~Kiros, K.~Cho, A.~Courville, R.~Salakhudinov, R.~Zemel, and
  Y.~Bengio, ``Show, attend and tell: Neural image caption generation with
  visual attention,'' in \emph{ICML}, 2015, pp. 2048--2057.

\bibitem{CVPR16:SA}
Q.~You, H.~Jin, Z.~Wang, C.~Fang, and J.~Luo, ``Image captioning with semantic
  attention,'' in \emph{CVPR}, 2016, pp. 4651--4659.

\bibitem{CVPR17:SCA}
L.~Chen, H.~Zhang, J.~Xiao, L.~Nie, J.~Shao, W.~Liu, and T.-S. Chua, ``Sca-cnn:
  Spatial and channel-wise attention in convolutional networks for image
  captioning,'' in \emph{CVPR}, 2017, pp. 5659--5667.

\bibitem{CVPR17:BA}
T.~Yao, Y.~Pan, Y.~Li, Z.~Qiu, and T.~Mei, ``Boosting image captioning with
  attributes,'' in \emph{CVPR}, 2017, pp. 4894--4902.

\bibitem{CVPR18:BUTD}
P.~Anderson, X.~He, C.~Buehler, D.~Teney, M.~Johnson, S.~Gould, and L.~Zhang,
  ``Bottom-up and top-down attention for image captioning and visual question
  answering,'' in \emph{CVPR}, 2018, pp. 6077--6086.

\bibitem{CVPR19:LBPF}
Y.~Qin, J.~Du, Y.~Zhang, and H.~Lu, ``Look back and predict forward in image
  captioning,'' in \emph{CVPR}, 2019, pp. 8367--8375.

\bibitem{ICCV19:AOA}
L.~Huang, W.~Wang, J.~Chen, and X.-Y. Wei, ``Attention on attention for image
  captioning,'' in \emph{ICCV}, 2019.

\bibitem{ICCV19:HIP}
T.~Yao, Y.~Pan, Y.~Li, and T.~Mei, ``Hierarchy parsing for image captioning,''
  in \emph{ICCV}, 2019.

\bibitem{ACL02:BLEU}
K.~Papineni, S.~Roukos, T.~Ward, and W.-J. Zhu, ``Bleu: a method for automatic
  evaluation of machine translation,'' in \emph{ACL}, 2002, pp. 311--318.

\bibitem{ACL05:METEOR}
S.~Banerjee and A.~Lavie, ``Meteor: An automatic metric for mt evaluation with
  improved correlation with human judgments,'' in \emph{ACL workshop}, 2005,
  pp. 65--72.

\bibitem{ACL04:ROUGE}
C.-Y. Lin, ``Rouge: A package for automatic evaluation of summaries,'' in
  \emph{ACL workshop}, 2004, pp. 74--81.

\bibitem{CVPR15:CIDER}
R.~Vedantam, C.~Lawrence~Zitnick, and D.~Parikh, ``Cider: Consensus-based image
  description evaluation,'' in \emph{CVPR}, 2015, pp. 4566--4575.

\bibitem{JAIR:FLICKR8K}
M.~Hodosh, P.~Young, and J.~Hockenmaier, ``Framing image description as a
  ranking task: Data, models and evaluation metrics,'' \emph{Journal of
  Artificial Intelligence Research}, vol.~47, pp. 853--899, 2013.

\bibitem{ACL14:FLICKR30K}
P.~Young, A.~Lai, M.~Hodosh, and J.~Hockenmaier, ``From image descriptions to
  visual denotations: New similarity metrics for semantic inference over event
  descriptions,'' \emph{Trans. of the Association for Computational
  Linguistics}, vol.~2, pp. 67--78, 2014.

\bibitem{ECCV14:COCO}
T.-Y. Lin, M.~Maire, S.~Belongie, J.~Hays, P.~Perona, D.~Ramanan,
  P.~Doll{\'a}r, and C.~L. Zitnick, ``Microsoft coco: Common objects in
  context,'' in \emph{ECCV}, 2014, pp. 740--755.

\bibitem{CVPR15:IMCAP}
O.~Vinyals, A.~Toshev, S.~Bengio, and D.~Erhan, ``Show and tell: A neural image
  caption generator,'' in \emph{CVPR}, 2015, pp. 3156--3164.

\bibitem{CVPR18:CONVIMCAP}
J.~Aneja, A.~Deshpande, and A.~G. Schwing, ``Convolutional image captioning,''
  in \emph{CVPR}, 2018, pp. 5561--5570.

\bibitem{CVPR17:SPIDER}
S.~Liu, Z.~Zhu, N.~Ye, S.~Guadarrama, and K.~Murphy, ``Improved image
  captioning via policy gradient optimization of spider,'' in \emph{CVPR},
  2017, pp. 873--881.

\bibitem{CVPR18:EIC}
Y.~Cui, G.~Yang, A.~Veit, X.~Huang, and S.~Belongie, ``Learning to evaluate
  image captioning,'' in \emph{CVPR}, 2018, pp. 5804--5812.

\bibitem{TMM:DRP}
W.~Xu, J.~Yu, Z.~Miao, L.~Wan, Y.~Tian, and Q.~Ji, ``Deep reinforcement
  polishing network for video captioning,'' \emph{IEEE Trans. on Multimedia},
  2020.

\bibitem{TMM:STAT}
C.~Yan, Y.~Tu, X.~Wang, Y.~Zhang, X.~Hao, Y.~Zhang, and Q.~Dai, ``Stat:
  spatial-temporal attention mechanism for video captioning,'' \emph{IEEE
  trans. on multimedia}, vol.~22, no.~1, pp. 229--241, 2019.

\bibitem{CVPR17:PSPNET}
H.~Zhao, J.~Shi, X.~Qi, X.~Wang, and J.~Jia, ``Pyramid scene parsing network,''
  in \emph{CVPR}, 2017, pp. 2881--2890.

\bibitem{BMVC18:PAN}
H.~Li, P.~Xiong, J.~An, and L.~Wang, ``Pyramid attention network for semantic
  segmentation,'' in \emph{BMVC}, 2018.

\bibitem{CVPR17:FPN}
T.-Y. Lin, P.~Doll{\'a}r, R.~Girshick, K.~He, B.~Hariharan, and S.~Belongie,
  ``Feature pyramid networks for object detection,'' in \emph{CVPR}, 2017, pp.
  2117--2125.

\bibitem{CVPR18:SE}
J.~Hu, L.~Shen, and G.~Sun, ``Squeeze-and-excitation networks,'' in
  \emph{CVPR}, 2018, pp. 7132--7141.

\bibitem{CVPR19:DAN}
J.~Fu, J.~Liu, H.~Tian, Y.~Li, Y.~Bao, Z.~Fang, and H.~Lu, ``Dual attention
  network for scene segmentation,'' in \emph{CVPR}, 2019, pp. 3146--3154.

\bibitem{TMM:VIDCAP}
L.~Gao, Z.~Guo, H.~Zhang, X.~Xu, and H.~T. Shen, ``Video captioning with
  attention-based lstm and semantic consistency,'' \emph{IEEE Trans. on
  Multimedia}, vol.~19, no.~9, pp. 2045--2055, 2017.

\bibitem{AAAI19:PSAC}
X.~Li, J.~Song, L.~Gao, X.~Liu, W.~Huang, X.~He, and C.~Gan, ``Beyond rnns:
  Positional self-attention with co-attention for video question answering,''
  in \emph{AAAI}, 2019, pp. 8658--8665.

\bibitem{MM17:AAC}
Y.~Bin, Y.~Yang, J.~Zhou, Z.~Huang, and H.~T. Shen, ``Adaptively attending to
  visual attributes and linguistic knowledge for captioning,'' in \emph{ACM
  MM}, 2017, pp. 1345--1353.

\bibitem{NIPS17:TRANSFORMER}
A.~Vaswani, N.~Shazeer, N.~Parmar, J.~Uszkoreit, L.~Jones, A.~N. Gomez,
  {\L}.~Kaiser, and I.~Polosukhin, ``Attention is all you need,'' in
  \emph{NIPS}, 2017, pp. 5998--6008.

\bibitem{NC:LSTM}
S.~Hochreiter and J.~Schmidhuber, ``Long short-term memory,'' \emph{Neural
  computation}, vol.~9, no.~8, pp. 1735--1780, 1997.

\bibitem{CVPR17:VFS}
L.~Mai, H.~Jin, Z.~Lin, C.~Fang, J.~Brandt, and F.~Liu, ``Spatial-semantic
  image search by visual feature synthesis,'' in \emph{CVPR}, 2017, pp.
  4718--4727.

\bibitem{AAAI:DESAN}
T.~Shen, T.~Zhou, G.~Long, J.~Jiang, S.~Pan, and C.~Zhang, ``Disan: Directional
  self-attention network for rnn/cnn-free language understanding,'' in
  \emph{AAAI}, 2018.

\bibitem{ARXIV:RESNEST}
H.~Zhang, C.~Wu, Z.~Zhang, Y.~Zhu, Z.~Zhang, H.~Lin, Y.~Sun, T.~He, J.~Muller,
  R.~Manmatha, M.~Li, and A.~Smola, ``Resnest: Split-attention networks,''
  \emph{arXiv preprint arXiv:2004.08955}, 2020.

\bibitem{CVPR16:RESNET}
K.~He, X.~Zhang, S.~Ren, and J.~Sun, ``Deep residual learning for image
  recognition,'' in \emph{CVPR}, 2016, pp. 770--778.

\bibitem{CVPR17:SCSIC}
S.~J. Rennie, E.~Marcheret, Y.~Mroueh, J.~Ross, and V.~Goel, ``Self-critical
  sequence training for image captioning,'' in \emph{CVPR}, 2017, pp.
  7008--7024.

\bibitem{CVPR15:KARPATHY}
A.~Karpathy and L.~Fei-Fei, ``Deep visual-semantic alignments for generating
  image descriptions,'' in \emph{CVPR}, 2015, pp. 3128--3137.

\bibitem{ICLR15:VGG}
K.~Simonyan and A.~Zisserman, ``Very deep convolutional networks for
  large-scale image recognition,'' in \emph{ICLR}, 2015.

\bibitem{ICLR19:ADAMW}
I.~Loshchilov and F.~Hutter, ``Decoupled weight decay regularization,'' in
  \emph{ICLR}, 2019.

\bibitem{CVPR18:DODC}
R.~Luo, B.~Price, S.~Cohen, and G.~Shakhnarovich, ``Discriminability objective
  for training descriptive captions,'' in \emph{CVPR}, 2018, pp. 6964--6974.

\bibitem{MM16:BLSTM}
C.~Wang, H.~Yang, C.~Bartz, and C.~Meinel, ``Image captioning with deep
  bidirectional lstms,'' in \emph{MM}, 2016, pp. 988--997.

\bibitem{TOMM:AS}
M.~Cornia, L.~Baraldi, G.~Serra, and R.~Cucchiara, ``Paying more attention to
  saliency: Image captioning with saliency and context attention,'' \emph{ACM
  Trans. on Multimedia Computing, Communications, and Applications}, vol.~14,
  no.~2, pp. 1--21, 2018.

\bibitem{AAAI17:AC}
C.~Liu, J.~Mao, F.~Sha, and A.~Yuille, ``Attention correctness in neural image
  captioning,'' in \emph{AAAI}, 2017.

\bibitem{ICCV17:LCNN}
J.~Gu, G.~Wang, J.~Cai, and T.~Chen, ``An empirical study of language cnn for
  image captioning,'' in \emph{ICCV}, 2017, pp. 1222--1231.

\bibitem{CVPR17:ADA}
J.~Lu, C.~Xiong, D.~Parikh, and R.~Socher, ``Knowing when to look: Adaptive
  attention via a visual sentinel for image captioning,'' in \emph{CVPR}, 2017,
  pp. 375--383.

\bibitem{TMM:STP}
L.~Guo, J.~Liu, S.~Lu, and H.~Lu, ``Show, tell and polish: Ruminant decoding
  for image captioning,'' \emph{IEEE Trans. on Multimedia}, vol.~22, no.~8, pp.
  2149--2162, 2019.

\bibitem{TPAMI:HLSTMAT}
L.~Gao, X.~Li, J.~Song, and H.~T. Shen, ``Hierarchical lstms with adaptive
  attention for visual captioning,'' \emph{IEEE trans. on pattern analysis and
  machine intelligence}, 2019.

\bibitem{TMM:SEMANTIC}
Z.~Zhang, Q.~Wu, Y.~Wang, and F.~Chen, ``High-quality image captioning with
  fine-grained and semantic-guided visual attention,'' \emph{IEEE Trans. on
  Multimedia}, vol.~21, no.~7, pp. 1681--1693, 2018.

\bibitem{TMM:IMCAP_SG}
X.~Li and S.~Jiang, ``Know more say less: Image captioning based on scene
  graphs,'' \emph{IEEE Trans. on Multimedia}, vol.~21, no.~8, pp. 2117--2130,
  2019.

\bibitem{ECCV18:GCN_LSTM}
T.~Yao, Y.~Pan, Y.~Li, and T.~Mei, ``Exploring visual relationship for image
  captioning,'' in \emph{ECCV}, 2018, pp. 684--699.

\bibitem{NIPS19:AAT}
L.~Huang, W.~Wang, Y.~Xia, and J.~Chen, ``Adaptively aligned image captioning
  via adaptive attention time,'' in \emph{NIPS}, 2019, pp. 8942--8951.

\bibitem{TMM:GLD}
J.~Wu, T.~Chen, H.~Wu, Z.~Yang, G.~Luo, and L.~Lin, ``Fine-grained image
  captioning with global-local discriminative objective,'' \emph{IEEE Trans. on
  Multimedia}, 2020.

\bibitem{MM18:VQA}
L.~Gao, P.~Zeng, J.~Song, X.~Liu, and H.~T. Shen, ``Examine before you answer:
  Multi-task learning with adaptive-attentions for multiple-choice vqa,'' in
  \emph{MM}, 2018, pp. 1742--1750.

\bibitem{MM17:MMA}
J.~Xu, T.~Yao, Y.~Zhang, and T.~Mei, ``Learning multimodal attention lstm
  networks for video captioning,'' in \emph{MM}, 2017, pp. 537--545.

\end{thebibliography}

\begin{IEEEbiography}[{\includegraphics[width=1in,height=1.25in,clip,keepaspectratio]{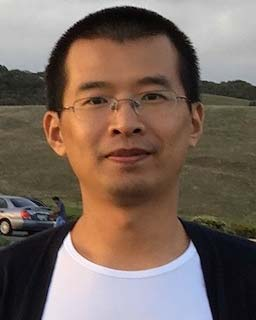}}]{Litao Yu}
received the Ph.D. degree from The University of Queensland, Brisbane, QLD, Australia, in 2016. He then worked as a Research Fellow with the QUT Centre of Robotics, Queensland University of Technology, Brisbane, in 2017. He is currently a Research Fellow with the Multimedia Data Analytics Lab (MDAL), Global Big Data Technologies Centre (GBDTC), University of Technology Sydney, Ultimo, NSW, Australia. He is also an Adjunct Research Fellow with the Institute for Integrated and Intelligent Systems, Griffith University, Nathan, QLD, Australia. His research interests include machine learning, multimedia content analysis, and image processing.
\end{IEEEbiography}

\begin{IEEEbiography}[{\includegraphics[width=1in,height=1.25in,clip,keepaspectratio]{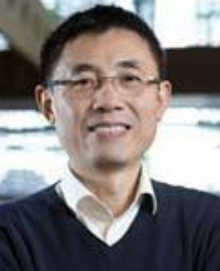}}]{Jian Zhang} 
(Senior Member, IEEE) is currently an Associate Professor with the Faculty of Engineering and IT, University of Technology Sydney (UTS), Ultimo, NSW, Australia. He is also the Director of the Multimedia Data Analytics Lab (MDAL), Global Big Data Technologies Centre (GBDTC). He has actively engaged in research collaboration with industry labs, supervised Ph.D. research students, and developed new multimedia analytics courses at UTS. Since 2011, as a leading chief investigator at the Faculty of Engineering and IT, he has led more than ten research projects with industry labs, including Microsoft Research, Nokia Research Centre, Tampere, Finland, Toshiba (Australia) R/D, Sydney, NSW, Australia, and Canon, Tokyo, Japan. He has coauthored more than 180 papers in top journals and refereed conference proceedings from his research output and filed more than ten patents in USA, U.K., and Australia, including six U.S.-issued patents. His research interests include 2-D- and 3-D-based computer vision, pattern recognition and data analytics, large-scale image and video content analytics, retrieval and mining, and multimedia and social media signal processing.
\end{IEEEbiography}


\begin{IEEEbiography}[{\includegraphics[width=1in,height=1.25in,clip,keepaspectratio]{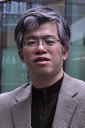}}]{Qiang Wu}
(Senior Member, IEEE) received the Ph.D. degree from the University of Technology Sydney, Ultimo, NSW, Australia, in 2004. He is currently an Associate Professor and a Core Member of the Global Big Data Technologies Center, University of Technology Sydney. His research interests include computer vision, image processing, pattern recognition, machine learning, and multimedia processing. He serves as a Reviewer for several journals, including TPAMI, TIP, TCSVT, and TSMCB. His research has been published in many premier international conferences including ECCV, CVPR, ICIP, and ICPR, and major international journals, such as TIP, TSMCB, TCSVT and TSP.
\end{IEEEbiography}

\end{document}